\documentclass[10pt,twocolumn,letterpaper]{article}

\PassOptionsToPackage{bookmarks={false}}{hyperref}
\usepackage{wacv}
\usepackage{times}
\usepackage{epsfig}
\usepackage{graphicx}
\usepackage{hyperref}
\usepackage{url}
\usepackage{amsmath}
\usepackage{amssymb}
\usepackage{amsfonts}
\usepackage{booktabs}
\usepackage{siunitx}
\usepackage{soul}
\usepackage{authblk}

\wacvfinalcopy % *** Uncomment this line for the final submission

\def\wacvPaperID{104} % *** Enter the wacv Paper ID here

\ifwacvfinal\fi
\begin{document}

%%%%%%%%% TITLE
\title{Understanding Convolution for Semantic Segmentation}

\author[Wang et al.]
       {Panqu Wang$^1$, Pengfei Chen$^1$, Ye Yuan$^{2}$,
       Ding Liu$^3$, Zehua Huang$^1$, Xiaodi Hou$^1$, 
       Garrison Cottrell$^4$\\
       $^1$TuSimple, 
       $^2$Carnegie Mellon University, 
       $^3$University of Illinois Urbana-Champaign, 
       $^4$UC San Diego\\
       \textit{$^1$\{panqu.wang,pengfei.chen,zehua.huang,xiaodi.hou\}@tusimple.ai,
       	       $^2$yey1@andrew.cmu.edu,
               $^3$dingliu2@illinois.edu,
               $^4$gary@ucsd.edu} 
       }
% Authors at the same institution
%\author{First Author \hspace{2cm} Second Author \\
% Authors at different institutions
% \author{Panqu Wang\\
% TuSimple\\
% {\tt\small panqu.wang@tusimple.ai}
% \and
% Pengfei Chen\\
% TuSimple\\
% {\tt\small pengfei.chen@tusimple.ai}
% \and
% Ye Yuan\\
% Carnegie Mellon University\\
% {\tt\small yey1@andrew.cmu.edu}
% \and
% Ding Liu\\
% UIUC\\
% \and
% Zehua Huang\\
% TuSimple\\
% \and
% Xiaodi Hou\\
% TuSimple\\
% \and
% Garrison Cottrell\\
% UC San Diego\\
% }
\maketitle
\ifwacvfinal\thispagestyle{empty}\fi

%%%%%%%%% ABSTRACT
\begin{abstract}
Recent advances in deep learning, especially deep convolutional neural networks (CNNs), have led to significant improvement over previous semantic segmentation systems. Here we show how to improve pixel-wise semantic segmentation by manipulating convolution-related operations that are of both theoretical and practical value. First, we design \textit{dense upsampling convolution (DUC)} to generate pixel-level prediction, which is able to capture and decode more detailed information that is generally missing in bilinear upsampling. Second, we propose a \textit{hybrid dilated convolution (HDC)} framework in the encoding phase. This framework 1) effectively enlarges the receptive fields (RF) of the network to aggregate global information; 2) alleviates what we call the \lq\lq gridding issue\rq\rq caused by the standard dilated convolution operation. We evaluate our approaches thoroughly on the Cityscapes dataset, and achieve a state-of-art result of 80.1\% mIOU in the test set at the time of submission. We also have achieved state-of-the-art overall on the KITTI road estimation benchmark and the PASCAL VOC2012 segmentation task. Our source code can be found at \href{https://github.com/TuSimple/TuSimple-DUC}{\normalfont{\url{https://github.com/TuSimple/TuSimple-DUC}}} .
\end{abstract}

%%%%%%%%% BODY TEXT
\section{Introduction}

Semantic segmentation aims to assign a categorical label to every pixel in an image, which plays an important role in image understanding and self-driving systems. The recent success of deep convolutional neural network (CNN) models \cite{krizhevsky2012imagenet,simonyan2014very,he2015deep} has enabled remarkable progress in pixel-wise semantic segmentation tasks due to rich hierarchical features and an end-to-end trainable framework \cite{long2015fully,zheng2015conditional,yu2015multi,liu2015semantic,lin2015efficient,chen2016deeplab}. Most state-of-the-art semantic segmentation systems have three key components:1) a fully-convolutional network (FCN), first introduced in \cite{long2015fully}, replacing the last few fully connected layers by convolutional layers to make efficient end-to-end learning and inference that can take arbitrary input size; 2) Conditional Random Fields (CRFs), to capture both local and long-range dependencies within an image to refine the prediction map; 3) dilated convolution (or Atrous convolution), which is used to increase the resolution of intermediate feature maps in order to generate more accurate predictions while maintaining the same computational cost. 

Since the introduction of FCN in \cite{long2015fully}, improvements on fully-supervised semantic segmentation systems are generally focused on two perspectives: First, applying deeper FCN models. Significant gains in mean Intersection-over-Union (mIoU) scores on PASCAL VOC2012 dataset \cite{pascal-voc-2012} were reported when the 16-layer VGG-16 model \cite{simonyan2014very} was replaced by a 101-layer ResNet-101 \cite{he2015deep} model \cite{chen2016deeplab}; using 152 layer ResNet-152 model yields further improvements \cite{wu2016high}. This trend is consistent with the performance of these models on ILSVRC \cite{russakovsky2015imagenet} object classification tasks, as deeper networks generally can model more complex representations and learn more discriminative features that better distinguish among categories. Second, making CRFs more powerful. This includes applying fully connected pairwise CRFs \cite{koltun2011efficient} as a post-processing step \cite{chen2016deeplab}, integrating CRFs into the network by approximating its mean-field inference steps \cite{zheng2015conditional,liu2015semantic,lin2015efficient} to enable end-to-end training, and incorporating additional information into CRFs such as edges \cite{kokkinos2015pushing} and object detections \cite{arnab2015higher}.

We are pursuing further improvements on semantic segmentation from another perspective: the \textit{convolutional} operations for both decoding (from intermediate feature map to output label map) and encoding (from input image to feature map) counterparts. In decoding, most state-of-the-art semantic segmentation systems simply use bilinear upsampling (before the CRF stage) to get the output label map \cite{lin2015efficient,liu2015semantic,chen2016deeplab}. Bilinear upsampling is not learnable and may lose fine details. Inspired by work in image super-resolution \cite{shi2016real}, we propose a method called  \textit{dense upsampling convolution (DUC)}, which is extremely easy to implement and can achieve pixel-level accuracy: instead of trying to recover the full-resolution label map at once, we learn an array of upscaling filters to upscale the downsized feature maps into the final dense feature map of the desired size. DUC naturally fits the FCN framework by enabling end-to-end training, and it increases the mIOU of pixel-level semantic segmentation on the Cityscapes dataset \cite{Cordts2016Cityscapes} significantly, especially on objects that are relatively small. 

For the encoding part, dilated convolution recently became popular \cite{chen2016deeplab,yu2015multi,wu2016high,zhou2016semantic}, as it maintains the resolution and receptive field of the network by in inserting \lq\lq holes\rq\rq in the convolution kernels, thus eliminating the need for downsampling (by max-pooling or strided convolution). However, an inherent problem exists in the current dilated convolution framework, which we identify as \lq\lq gridding\rq\rq: as zeros are padded between two pixels in a convolutional kernel, the receptive field of this kernel only covers an area with checkerboard patterns - only locations with non-zero values are sampled, losing some neighboring information. The problem gets worse when the rate of dilation increases, generally in higher layers when the receptive field is large: the convolutional kernel is too sparse to cover any local information, since the non-zero values are too far apart. Information that contributes to a fixed pixel always comes from its predefined gridding pattern, thus losing a huge portion of information. Here we propose a simple \textit{hybrid dilation convolution (HDC)} framework as a first attempt to address this problem: instead of using the same rate of dilation for the same spatial resolution, we use a range of dilation rates and concatenate them serially the same way as \lq\lq blocks\rq\rq in ResNet-101 \cite{he2015deep}. We show that HDC helps the network to alleviate the gridding problem. Moreover, choosing proper rates can effectively increases the receptive field size and improves the accuracy for objects that are relatively big.

We design DUC and HDC to make \textit{convolution} operations better serve the need of pixel-level semantic segmentation. The technical details are described in Section 3 below. Combined with post-processing by Conditional Random Fields (CRFs), we show that this approach achieves state-of-the art performance on the Cityscapes pixel-level semantic labeling task, KITTI road estimation benchmark, and PASCAL VOC2012 segmentation task.

\section{Related Work}
\textbf{Decoding of Feature Representation:} In the pixel-wise semantic segmentation task, the output label map has the same size as the input image. Because of the operation of max-pooling or strided convolution in CNNs, the size of feature maps of the last few layers of the network are inevitably downsampled. Multiple approaches have been proposed to decode accurate information from the downsampled feature map to label maps. Bilinear interpolation is commonly used \cite{lin2015efficient,liu2015semantic,chen2016deeplab}, as it is fast and memory-efficient. Another popular method is called deconvolution, in which the unpooling operation, using stored pooling switches from the pooling step, recovers the information necessary for feature visualization \cite{zeiler2014visualizing}. In \cite{long2015fully}, a single deconvolutional layer is added in the decoding stage to produce the prediction result using stacked feature maps from intermediate layers. In \cite{dosovitskiy2015learning}, multiple deconvolutional layers are applied to generate chairs, tables, or cars from several attributes. Noh et al. \cite{noh2015learning} employ deconvolutional layers as mirrored version of convolutional layers by using stored pooled location in unpooling step. \cite{noh2015learning} show that coarse-to-fine object structures, which are crucial to recover fine-detailed information, can be reconstructed along the propagation of the deconvolutional layers. Fischer at al. \cite{fischer2015flownet} use a similar mirrored structure, but combine information from multiple deconvolutional layers and perform upsampling to make the final prediction.

\textbf{Dilated Convolution:} Dilated Convolution (or Atrous convolution) was originally developed in \textit{algorithme \`{a} trous} for wavelet decomposition \cite{holschneider1990real}. The main idea of dilated convolution is to insert \lq\lq holes\rq\rq (zeros) between pixels in convolutional kernels to increase image resolution, thus enabling dense feature extraction in deep CNNs. In the semantic segmentation framework, dilated convolution is also used to enlarge the field of convolutional kernels. Yu \& Koltun \cite{yu2015multi} use serialized layers with increasing rates of dilation to enable context aggregation, while \cite{chen2016deeplab} design an \lq\lq atrous spatial pyramid pooling (ASPP)\rq\rq scheme to capture multi-scale objects and context information by placing multiple dilated convolution layers in parallel. More recently, dilated convolution has been applied to a broader range of tasks, such as object detection \cite{dai2016r}, optical flow \cite{sevilla2016optical}, and audio generation \cite{wavenet}.

\section{Our Approach}
\subsection{Dense Upsampling Convolution (DUC)}
\begin{figure*}[t]
\begin{center}
\includegraphics[width=0.99\textwidth]{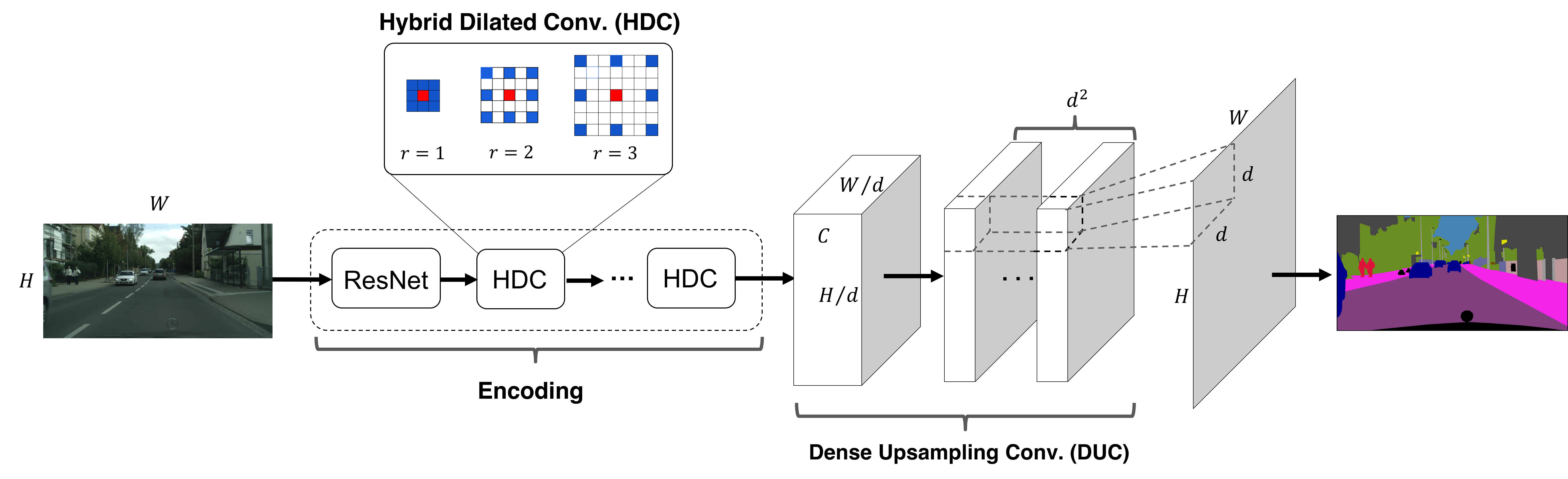}
\end{center}
\caption{Illustration of the architecture of ResNet-101 network with Hybrid Dilated Convolution (HDC) and Dense Upsampling Convolution (DUC) layer. HDC is applied within ResNet blocks, and DUC is applied on top of network and is used for decoding purpose.}
\label{Figure1_DUC}
\vspace{-1pt}
\end{figure*}

Suppose an input image has height $H$, width $W$, and color channels $C$, and the goal of pixel-level semantic segmentation is to generate a label map with size $H\times{W}$ where each pixel is labeled with a category label. After feeding the image into a deep FCN, a feature map with dimension $h\times{w}\times{c}$ is obtained at the final layer before making predictions, where $h=H/d$, $w=W/d$, and $d$ is the downsampling factor. Instead of performing bilinear upsampling, which is not learnable, or using deconvolution network (as in \cite{noh2015learning}), in which zeros have to be padded in the unpooling step before the convolution operation, DUC applies convolutional operations directly on the feature maps to get the dense pixel-wise prediction map. Figure~\ref{Figure1_DUC} depicts the architecture of our network with a DUC layer.

The DUC operation is all about \textit{convolution}, which is performed on the feature map from ResNet of dimension $h\times{w}\times{c}$ to get the output feature map of dimension $h\times{w}\times{(d^2\times{L})}$, where $L$ is the total number of classes in the semantic segmentation task. Thus each layer of the dense convolution is learning the prediction for each pixel. The output feature map is then reshaped to $H\times{W}\times{L}$ with a softmax layer, and an elementwise \textit{argmax} operator is applied to get the final label map. In practice, the ``reshape'' operation may not be necessary, as the feature map can be collapsed directly to a vector to be fed into the softmax layer. The key idea of DUC is to divide the whole label map into equal $d^2$ subparts which have the same height and width as the incoming feature map. This is to say, we transform the whole label map into a smaller label map with multiple channels. This transformation allows us to apply the convolution operation directly between the input feature map and the output label maps without the need of inserting extra values in deconvolutional networks (the \lq\lq unpooling\rq\rq operation). 

Since DUC is learnable, it is capable of capturing and recovering fine-detailed information that is generally missing in the bilinear interpolation operation. For example, if a network has a downsample rate of $1/16$, and an object has a length or width less than 16 pixels (such as a pole or a person far away), then it is more than likely that bilinear upsampling will not be able to recover this object. Meanwhile, the corresponding training labels have to be downsampled to correspond with the output dimension, which will already cause information loss for fine details. The prediction of DUC, on the other hand, is performed at the original resolution, thus enabling pixel-level decoding. In addition, the DUC operation can be naturally integrated into the FCN framework, and makes the whole encoding and decoding process end-to-end trainable.

\subsection{Hybrid Dilated Convolution (HDC)}
In 1-D, dilated convolution is defined as:
\begin{equation}
g[i]=\sum_{l=1}^{L}f[i+r\cdot l]h[l],
\end{equation}
where $f[i]$ is the input signal, $g[i]$ is the output signal , $h[l]$ denotes the filter of length $L$, and $r$ corresponds to the dilation rate we use to sample $f[i]$. In standard convolution, $r=1$.

In a semantic segmentation system, 2-D dilated convolution is constructed by inserting \lq\lq holes\rq\rq (zeros) between each pixel in the convolutional kernel. For a convolution kernel with size $k\times{k}$, the size of resulting dilated filter is $k_d\times{k_d}$, where $k_d=k+(k-1)\cdot(r-1)$. Dilated convolution is used to maintain high resolution of feature maps in FCN through replacing the max-pooling operation or strided convolution layer while maintaining the receptive field (or \lq\lq field of view\rq\rq in \cite{chen2016deeplab}) of the corresponding layer. For example, if a convolution layer in ResNet-101 has a stride $s=2$, then the stride is reset to $1$ to remove downsampling, and the dilation rate $r$ is set to 2 for all convolution kernels of subsequent layers. This process is applied iteratively through all layers that have a downsampling operation, thus the feature map in the output layer can maintain the same resolution as the input layer. In practice, however, dilated convolution is generally applied on feature maps that are already downsampled to achieve a reasonable efficiency/accuracy trade-off \cite{chen2016deeplab}. 

\begin{figure}[tp]
\begin{center}
\includegraphics[width=0.4\textwidth]{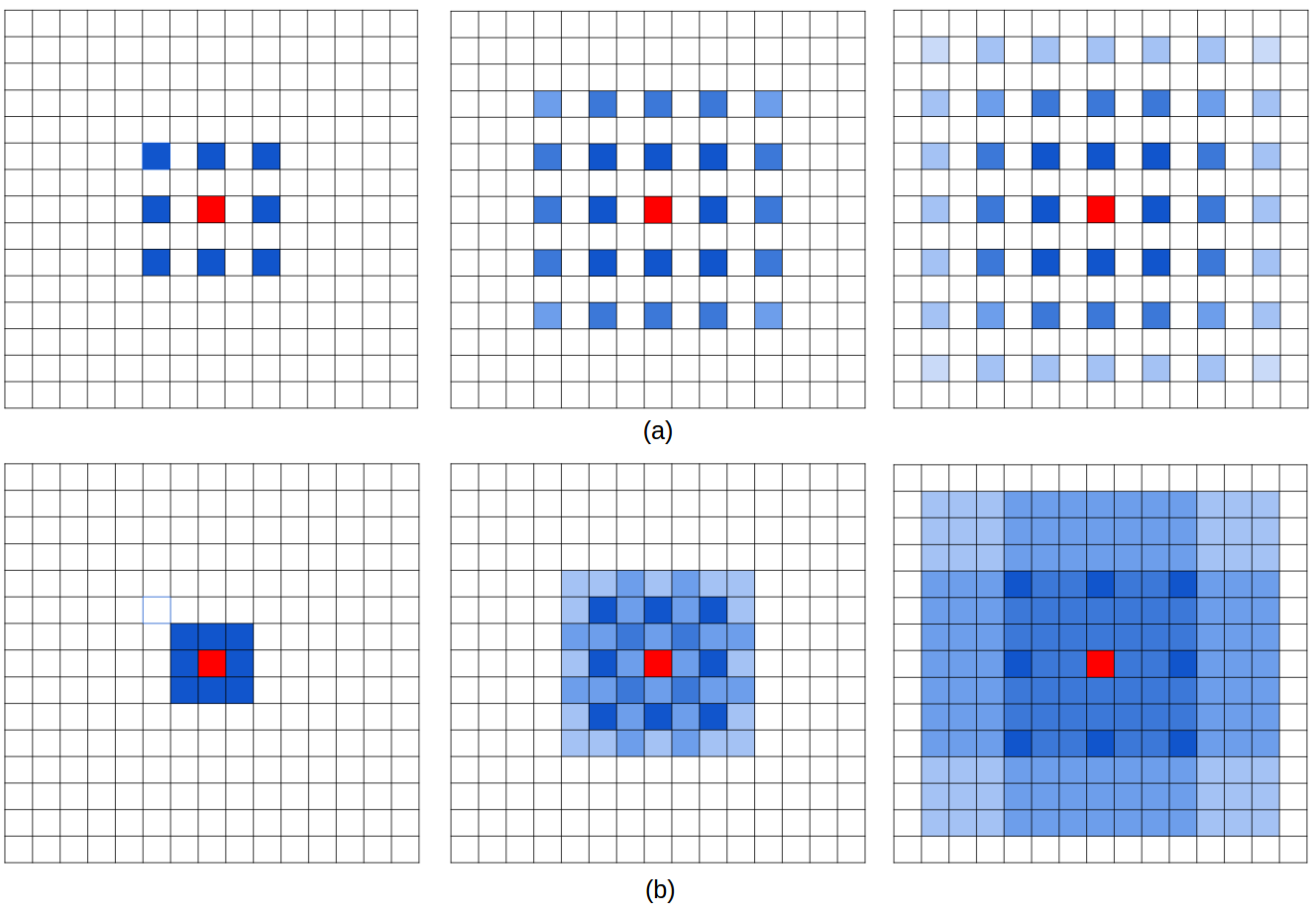}
\end{center}
\caption{Illustration of the gridding problem. Left to right: the pixels (marked in blue) contributes to the calculation of the center pixel (marked in red) through three convolution layers with kernel size $3\times 3$. (a) all convolutional layers have a dilation rate $r=2$. (b) subsequent convolutional layers have dilation rates of $r=1$, $2$, $3$, respectively.}
\vspace{-15pt}
\label{Figure2_gridding}
\end{figure}

However, one theoretical issue exists in the above dilated convolution framework, and we call it \lq\lq \textbf{gridding}\rq\rq (Figure~\ref{Figure2_gridding}):  For a pixel $p$ in a dilated convolutional layer $l$, the information that contributes to pixel $p$ comes from a nearby $k_d\times k_d$ region in layer $l-1$ centered at $p$. Since dilated convolution introduces zeros in the convolutional kernel, the actual pixels that participate in the computation from the $k_d\times k_d$ region are just $k\times k$, with a gap of $r-1$ between them. If $k=3, r=2$, only 9 out of 25 pixels in the region are used for the computation (Figure~\ref{Figure2_gridding} (a)). Since all layers have equal dilation rates $r$, then for pixel $p$ in the top dilated convolution layer $l_{top}$, the maximum possible number of locations that contribute to the calculation of the value of $p$ is $(w'\times h')/r^2$ where $w', h'$ are the width and height of the bottom dilated convolution layer, respectively. As a result, pixel $p$ can only view information in a checkerboard fashion, and lose a large portion (at least $75\%$ when $r=2$) of information. When $r$ becomes large in higher layers due to additional downsampling operations, the sample from the input can be very sparse, which may not be good for learning because 1) local information is completely missing; 2) the information can be irrelevant across large distances. Another outcome of the gridding effect is that pixels in nearby $r\times r$ regions at layer $l$ receive information from completely different set of \lq\lq grids\rq\rq， which may impair the consistency of local information.

Here we propose a simple solution- \textit{hybrid dilated convolution (HDC)}, to address this theoretical issue. Suppose we have $N$ convolutional layers with kernel size $K\times K$ that have dilation rates of $[r_1,...,r_i,...,r_n]$, the goal of HDC is to let the final size of the RF of a series of convolutional operations fully covers a square region without any holes or missing edges. We define the ``maximum distance between two nonzero values'' as
\begin{equation}
M_i=max[M_{i+1}-2r_i,M_{i+1}-2(M_{i+1}-r_i),r_i],
\end{equation}
with $M_n=r_n$. The design goal is to let $M_2\leq K$. %For example, $r_{i+1}=7$ (a kernel with 6 zeros in between like *000000* where * represents a nonzero value) followed by $r_i=2$ results in $M_i=3$ (*0*00*0*).
For example, for kernel size $K=3$, an $r=[1,2,5]$ pattern works as $M_2=2$; however, an $r=[1,2,9]$ pattern does not work as $M_2=5$. Practically, instead of using the same dilation rate for all layers after the downsampling occurs, we use a different dilation rate for each layer. In our network, the assignment of dilation rate follows a sawtooth wave-like heuristic: a number of layers are grouped together to form the \lq\lq rising edge\rq\rq of the wave that has an increasing dilation rate, and the next group repeats the same pattern. For example, for all layers that have dilation rate $r=2$, we form 3 succeeding layers as a group, and change their dilation rates to be 1, 2, and 3, respectively. By doing this, the top layer can access information from a broader range of pixels, in the same region as the original configuration (Figure~\ref{Figure2_gridding} (b)). This process is repeated through all layers, thus making the receptive field unchanged at the top layer.

Another benefit of HDC is that it can use arbitrary dilation rates through the process, thus naturally enlarging the receptive fields of the network without adding extra modules \cite{yu2015multi}, which is important for recognizing objects that are relatively big. One important thing to note, however, is that the dilation rate within a group should not have a common factor relationship (like 2,4,8, etc.), otherwise the gridding problem will still hold for the top layer. This is a key difference between our HDC approach and the atrous spatial pyramid pooling (ASPP) module in \cite{chen2016deeplab}, or the context aggregation module in \cite{yu2015multi}, where dilation factors that have common factor relationships are used. In addition, HDC is naturally integrated with the original layers of the network, without any need to add extra modules as in \cite{yu2015multi,chen2016deeplab}.

\section{Experiments and Results}
We report our experiments and results on three challenging semantic segmentation datasets: Cityscapes \cite{Cordts2016Cityscapes}, KITTI dataset \cite{Fritsch2013ITSC} for road estimation, and PASCAL VOC2012 \cite{pascal-voc-2012}. We use ResNet-101 or ResNet-152 networks that have been pretrained on the ImageNet dataset as a starting point for all of our models. The output layer contains the number of semantic categories to be classified depending on the dataset (including background, if applicable). We use the cross-entropy error at each pixel over the categories. This is then summed over all pixel locations of the output map, and we optimize this objective function using standard Stochastic Gradient Descent (SGD). We use MXNet \cite{chen2015mxnet} to train and evaluate all of our models on NVIDIA TITAN X GPUs.

\subsection{Cityscapes Dataset}
The Cityscapes Dataset is a large dataset that focuses on semantic understanding of urban street scenes. The dataset contains 5000 images with fine annotations across 50 cities, different seasons, varying scene layout and background. The dataset is annotated with 30 categories, of which 19 categories are included for training and evaluation (others are ignored). The training, validation, and test set contains 2975, 500, and 1525 fine images, respectively. An additional 20000 images with coarse (polygonal) annotations are also provided, but are only used for training.
\vspace{-5pt}

\subsubsection{Baseline Model}
We use the DeepLab-V2 \cite{chen2016deeplab} ResNet-101 framework to train our baseline model. Specifically, the network has a downsampling rate of 8, and dilated convolution with rate of 2 and 4 are applied to \textit{res4b} and \textit{res5b} blocks, respectively. An ASPP module with dilation rate of 6, 12, 18, and 24 is added on top of the network to extract multiscale context information. The prediction maps and training labels are downsampled by a factor of 8 compared to the size of original images, and bilinear upsampling is used to get the final prediction. Since the image size in the Cityscapes dataset is $1024\times 2048$, which is too big to fit in the GPU memory, we partition each image into twelve $800\times 800$ patches with partial overlapping, thus augmenting the training set to have $35700$ images. This data augmentation strategy is to make sure all regions in an image can be visited. This is an improvement over random cropping, in which nearby regions may be visited repeatedly.

We train the network using mini-batch SGD with patch size $544\times 544$ (randomly cropped from the $800\times 800$ patch) and batch size 12, using multiple GPUs. The initial learning rate is set to $2.5\times 10^{-4}$, and a \lq\lq \textit{poly}\rq\rq   learning rate (as in \cite{chen2016deeplab}) with $power=0.9$ is applied. Weight decay is set to $5\times 10^{-4}$, and momentum is $0.9$. The network is trained for 20 epochs and achieves mIoU of $72.3\%$ on the validation set.
\vspace{-5pt}

\subsubsection{Dense Upsampling Convolution (DUC)} 
\begin{figure*}[ht]
\begin{center}
\includegraphics[width=0.95\textwidth]{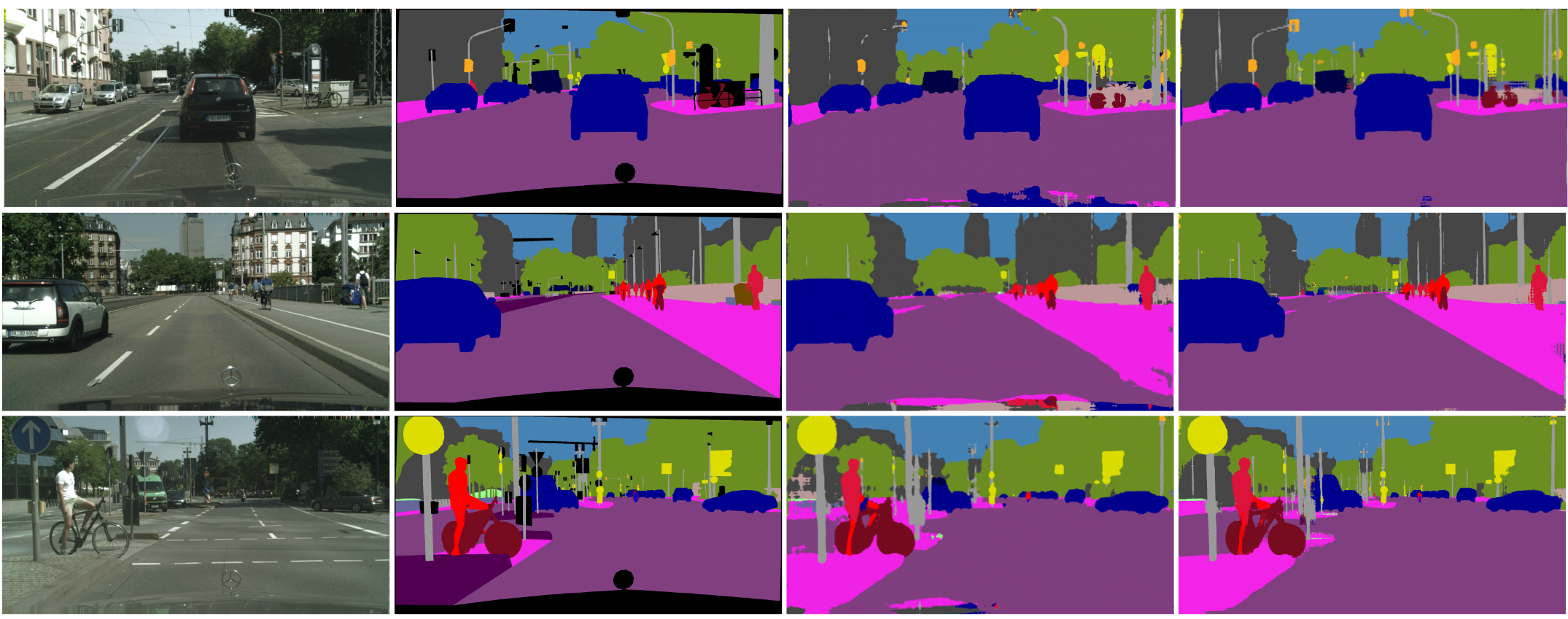}
\end{center}
\caption{Effect of Dense Upsampling Convolution (DUC) on the Cityscapes validation set. From left to right: input image, ground truth (areas with black color are ignored in evaluation), baseline model, and our ResNet-DUC model. }
\label{Figure3_vis_cityscapes_1}
\vspace{-15pt}
\end{figure*}

We examine the effect of DUC on the baseline network. In DUC， the only thing we change is the shape of the top convolutional layer. For example, if the dimension of the top convolutional layer is $68\times{68}\times{19}$ in the baseline model (19 is the number of classes), then the dimension of the same layer for a network with DUC will be $68\times{68}\times{(r^2\times 19)}$ where $r$ is the total downsampling rate of the network ($r=8$ in this case). The prediction map is then reshaped to size $544\times 544\times 19$. DUC will introduce extra parameters compared to the baseline model, but only at the top convolutional layer. We train the ResNet-DUC network the same way as the baseline model for 20 epochs, and achieve a mean IOU of \textbf{$74.3\%$} on the validation set, a $2\%$ increase compared to the baseline model. Visualization of the result of ResNet-DUC and comparison with the baseline model is shown in Figure~\ref{Figure3_vis_cityscapes_1}

From Figure~\ref{Figure3_vis_cityscapes_1}, we can clearly see that DUC is very helpful for identifying small objects, such as poles, traffic lights, and traffic signs. Consistent with our intuition, pixel-level dense upsampling can recover detailed information that is generally missed by bilinear interpolation.

\textbf{Ablation Studies}
We examine the effect of different settings of the network on the performance. Specifically, we examine: 1) the downsampling rate of the network, which controls the resolution of the intermediate feature map; 2) whether to apply the ASPP module, and the number of parallel paths in the module; 3) whether to perform 12-fold data augmentation; and 4) cell size, which determines the size of neighborhood region ($cell\times cell$) that one predicted pixel projects to. Pixel-level DUC should use  $cell=1$; however, since the ground-truth label generally cannot reach pixel-level precision, we also try $cell=2$ in the experiments. From Table~\ref{table_1}， we can see that making the downsampling rate smaller decreases the accuracy. Also it significantly raises the computational cost due to the increasing resolution of the feature maps. ASPP generally helps to improve the performance, and increasing ASPP channels from 4 to 6 (dilation rate 6 to 36 with interval 6) yields a $0.2\%$ boost. Data augmentation helps to achieve another $1.5\%$ improvement. Using $cell=2$ yields slightly better performance when compared with $cell=1$, and it helps to reduce computational cost by decreasing the channels of the last convolutional layer by a factor of 4.

\begin{table}[ht]
\begin{center}
\begin{tabular}{ccccc|c} \toprule[1.5pt]
{Network}      &     {DS}   & {ASPP}  &  {Augmentation} & {Cell} &{mIoU}\\ \midrule
   $Baseline$  &     $8$    &  $4$    &  $yes$  & $n/a$ & $72.3$ \\
   $Baseline$  &     $4$    &  $4$    &  $yes$  & $n/a$ & $70.9$ \\ \midrule
   $DUC$  &     	 $8$    &  $no$   &  $no$  &  $1$ & $71.9$ \\
   $DUC$  &    	     $8$    &  $4$    &  $no$  &  $1$ & $72.8$ \\
   $DUC$  &    	     $8$    &  $4$    &  $yes$  & $1$ & $74.3$ \\
   $DUC$  &     	 $4$    &  $4$    &  $yes$  & $1$ & $73.7$ \\ 
   $DUC$  &     	 $8$    &  $6$    &  $yes$  & $1$ & $74.5$ \\
   $DUC$  &     	 $8$    &  $6$    &  $yes$  & $2$ & $74.7$ \\ \bottomrule[1.5pt]
\end{tabular}
\vspace{3pt}
\caption{Ablation studies for applying ResNet-101 on the Cityscapes dataset. \textbf{DS}: Downsampling rate of the network.  \textbf{Cell}: neighborhood region that one predicted pixel represents.}
\label{table_1} 
\end{center}
\vspace{-15pt}
\end{table}

\textbf{Bigger Patch Size} Since setting $cell=2$ reduces GPU memory cost for network training, we explore the effect of patch size on the performance. Our assumption is that, since the original images are all $1024\times 2048$, the network should be trained using patches as big as possible in order to aggregate both local detail and global context information that may help learning. As such, we make the patch size to be $880\times 880$, and set the batch size to be 1 on each of the 4 GPUs used in training. Since the patch size exceeds the maximum dimension ($800\times 800$) in the previous 12-fold data augmentation framework, we adopt a new 7-fold data augmentation strategy: \iffalse \gary{Why can't you simply use 880-size patches in the same way?}\panqu{because the image is 1024x2048. Previously we use 2x6 configuration. Now since the patch is 880x880, so we can remove the height part.}\fi seven center locations with $x=512$, $y=\{256,512,...,1792\}$ are set in the original image; for each center location, a $880\times 880$ patch is obtained by randomly setting its center within a $160\times 160$ rectangle area centered at each center. This strategy makes sure that we can sample all areas in the image, including edges. Training with a bigger patch size boosts the performance to $\textbf{75.7\%}$, a $1\%$ improvement over the previous best result.

\textbf{Compared with Deconvolution}
We compare our DUC model with deconvolution, which also involves learning for upsampling. Particularly, we compare with 1) direct deconvolution from the prediction map (dowsampled by 8) to the original resolution; 2) deconvolution with an upsampling factor of 2 first, followed by an upsampling factor of 4. We design the deconv network to have approximately the same number of parameters as DUC. We use the ResNet-DUC bigger patch model to train the networks. The above two models achieve mIOU of $75.1\%$ and $75.0\%$, respectively, lower than the ResNet-DUC model ($75.7\%$ mIoU).

\textbf{Conditional Random Fields (CRFs)}
Fully-connected CRFs \cite{koltun2011efficient} are widely used for improving semantic segmentation quality as a post-processing step of an FCN \cite{chen2016deeplab}. We follow the formation of CRFs as shown in \cite{chen2016deeplab}. We perform a grid search on parameters on the validation set, and use $\sigma_{\alpha}=15$, $\sigma_{\beta}=3$, $\sigma_{\gamma}=1$ , $w_1=3$, and $w_2=3$ for all of our models. Applying CRFs to our best ResNet-DUC model yields an mIoU of $76.7\%$, a $1\%$ improvement over the model does not use CRFs.

\subsubsection{Hybrid Dilated Convolution (HDC)}
\begin{figure*}[h]
\begin{center}
\includegraphics[width=0.90\textwidth]{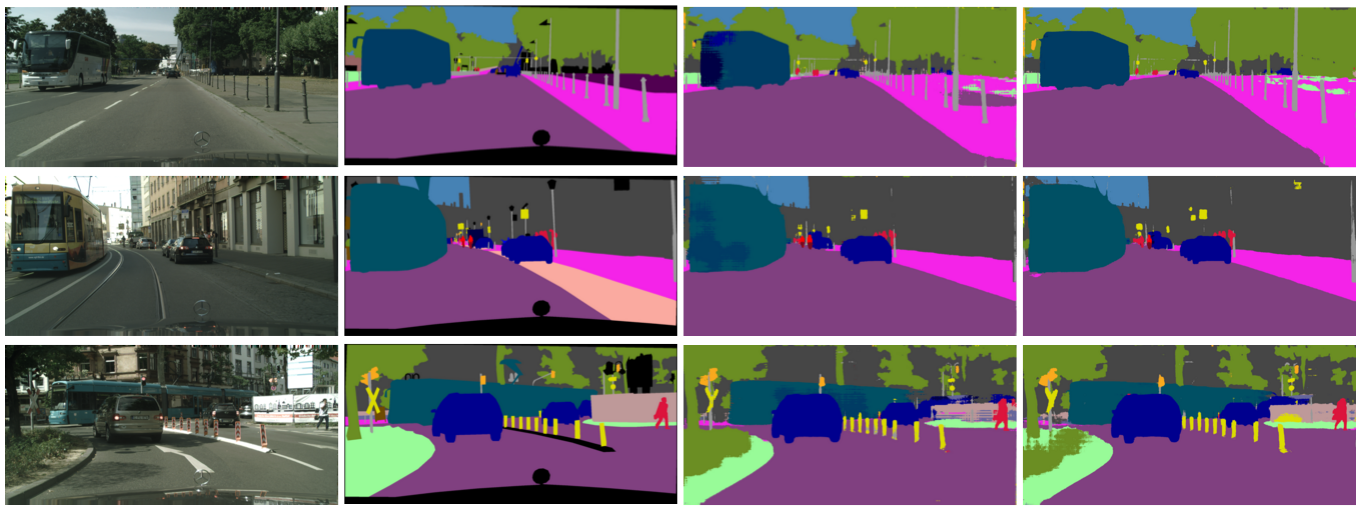}
\end{center}
\caption{Effect of Hybrid Dilated Convolution (HDC) on the Cityscapes validation set. From left to right: input image, ground truth, result of the ResNet-DUC model, result of the ResNet-DUC-HDC model (Dilation-bigger).}
\label{cvpr_2017_submission_figure_4}
\vspace{-15pt}
\end{figure*}

\begin{figure}[h]
\begin{center}
\includegraphics[width=0.45\textwidth]{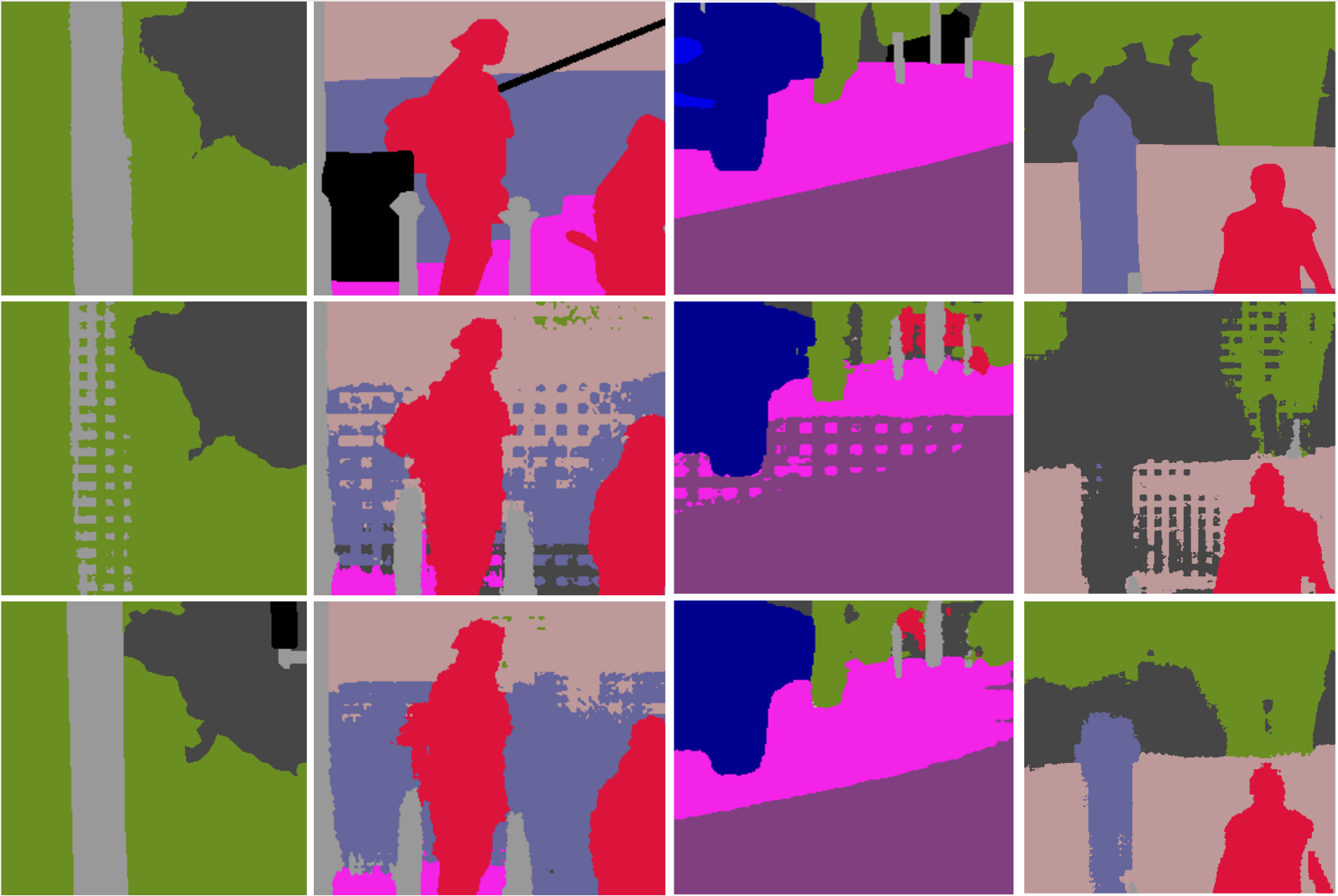}
\end{center}
\caption{Effectiveness of HDC in eliminating the gridding effect. First row: ground truth patch. Second row: prediction of the ResNet-DUC model. A strong gridding effect is observed. Third row: prediction of the ResNet-DUC-HDC (Dilation-RF) model.}
\label{Figure5_gridding_example_result}
\vspace{-10pt}
\end{figure}

We use the best 101 layer ResNet-DUC model as a starting point of applying HDC. Specifically, we experiment with several variants of the HDC module:
\begin{enumerate}
  \item No dilation: For all ResNet blocks containing dilation, we make their dilation rate $r=1$ (no dilation).
  \item Dilation-conv: For all blocks contain dilation, we group every 2 blocks together and make $r=2$ for the first block, and $r=1$ for the second block. 
  \item Dilation-RF: For the $res4b$ module that contains 23 blocks with dilation rate $r=2$, we group every 3 blocks together and change their dilation rates to be 1, 2, and 3, respectively. For the last two blocks, we keep $r=2$. For the $res5b$ module which contains 3 blocks with dilation rate $r=4$, we change them to 3, 4, and 5, respectively.
  \item Dilation-bigger: For $res4b$ module, we group every 4 blocks together and change their dilation rates to be 1, 2, 5, and 9, respectively. The rates for the last 3 blocks are 1, 2, and 5. For $res5b$ module, we set the dilation rates to be 5, 9, and 17.
\end{enumerate}
The result is summarized in Table~\ref{table_2}. We can see that increasing receptive field size generally yields higher accuracy. Figure~\ref{Figure5_gridding_example_result} illustrates the effectiveness of the ResNet-DUC-HDC model in eliminating the gridding effect. A visualization result is shown in Figure~\ref{cvpr_2017_submission_figure_4}. We can see our best ResNet-DUC-HDC model performs particularly well on objects that are relatively big. 
\begin{table}[ht]
\begin{center}
\begin{tabular}{cc|c} \toprule[1.75pt]
{Network}           &     {RF increased}  &{mIoU (without CRF)}\\ \midrule
   No dilation    &     $54$      & $72.9$\\
   Dilation-conv  &     $88$         & $75.0$ \\ 
   Dilation-RF    &     $116$      & $75.4$ \\
   Dilation-bigger &    $256$         & $76.4$ \\ \bottomrule[1.75pt]
\end{tabular}
\vspace{2pt}
\caption{Result of different variations of the HDC module. \lq\lq RF increased\rq\rq is the total size of receptive field increase along a single dimension compared to the layer before the dilation operation.}
\label{table_2} 
\end{center}
\vspace{-15pt}
\end{table}

\textbf{Deeper Networks}
We have also tried replacing our ResNet-101 based model with the ResNet-152 network, which is deeper and achieves better performance on the ILSVRC image classification task than ResNet-101 \cite{he2015deep}. Due to the network difference, we first train the ResNet-152 network to learn the parameters in all batch normalization (BN) layers for 10 epochs, and continue fine-tuning the network by fixing these BN parameters for another 20 epochs. The results are summarized in Table~\ref{table_3}. We can see that using the deeper ResNet-152 model generally yields better performance than the ResNet-101 model.
\begin{table}[ht]
\begin{center}
\begin{tabular}{ccc|c} \toprule[1.5pt]
{Network}      &     {Method} & {data}  &{mIoU}\\ \midrule
   ResNet-101  &     $Deconv$  & {fine}  & $75.1$\\
   ResNet-101  &     $DUC$+$HDC$ & {fine}   & $76.4$\\   
   ResNet-101  &     $DUC$+$HDC$  & {fine+coarse}  & $76.2$ \\\midrule
   ResNet-152  &     	 $Deconv$  & {fine}  & $76.4$ \\
   ResNet-152  &     	 $DUC$+$HDC$  & {fine}  & $76.7$\\
   ResNet-152  &     	 $DUC$+$HDC$  & {fine+coarse}  & $77.1$ \\
\bottomrule[1.5pt]
\end{tabular}
\vspace{2pt}
\caption{Effect of depth of the network and upsampling method for Cityscapes validation set (without CRF).}
\label{table_3} 
\end{center}
\vspace{-25pt}
\end{table}

\subsubsection{Test Set Results}
Our results on the Cityscapes test set are summarized in Table~\ref{table_4}. There are separate entries for models trained using fine-labels only, and using a combination of fine and coarse labels. Our ResNet-DUC-HDC model achieves $77.6\%$ mIoU using fine data only. Adding coarse data help us achieve $\textbf{78.5\%}$ mIoU.

In addition, inspired by the design of the VGG network \cite{simonyan2014very}, in that a single $5\times5$ convolutional layer can be decomposed into two adjacent $3\times3$ convolutional layers to increase the expressiveness of the network while maintaining the receptive field size, we replaced the $7\times 7$ convolutional layer in the original ResNet-101 network by three $3\times3$ convolutional layers. By retraining the updated network, we achieve a mIoU of $\textbf{80.1\%}$ on the test set using a single model without CRF post-processing. Our result achieves the state-of-the-art performance on the Cityscapes dataset at the time of submission. Compared with the strong baseline of Chen et al. \cite{chen2016deeplab}, we improve the mIoU by a significant margin ($9.7\%$), which demonstrates the effectiveness of our approach. 

\begin{table}[ht]
\begin{center}
\begin{tabular}{c|c} \toprule[1.5pt]
    {Method}             &{mIoU}\\ \midrule
    \textit{fine} & \\
   FCN 8s \cite{long2015fully}           & $65.3\%$\\
   Dilation10 \cite{yu2015multi}         & $67.1\%$ \\ 
   DeepLabv2-CRF \cite{chen2016deeplab}         & $70.4\%$ \\ 
   Adelaide\_context \cite{lin2015efficient}         & $71.6\%$ \\
   \textbf{ResNet-DUC-HDC (ours)}         & $\textbf{77.6\%}$ \\    \midrule[1.25pt]
   \textit{coarse} & \\ 
   LRR-4x \cite{ghiasi2016laplacian}           & $71.8\%$ \\
   {ResNet-DUC-HDC-Coarse }       & ${78.5\%}$\\
   \textbf{ResNet-DUC-HDC-Coarse (better network)}       & $\textbf{80.1\%}$\\ \bottomrule[1.5pt]

\end{tabular}
\vspace{2pt}
\caption{Performance on Cityscapes test set.}
\label{table_4} 
\vspace{-15pt}
\end{center}
\end{table}

\subsection{KITTI Road Segmentation}
\textbf{Dataset} The KITTI road segmentation task contains images of three various categories of road scenes, including 289 training images and 290 test images. The goal is to decide if each pixel in images is road or not. It is challenging to use neural network based methods due to the limited number of training images. In order to avoid overfitting, we crop patches of $320 \times 320$ pixels with a stride of $100$ pixels from the training images, and use the ResNet-101-DUC model pretrained from ImageNet during training. Other training settings are the same as Cityscapes experiment. We did not apply CRFs for post-processing.
\begin{figure}[h]
\begin{center}
\includegraphics[width=0.49\textwidth]{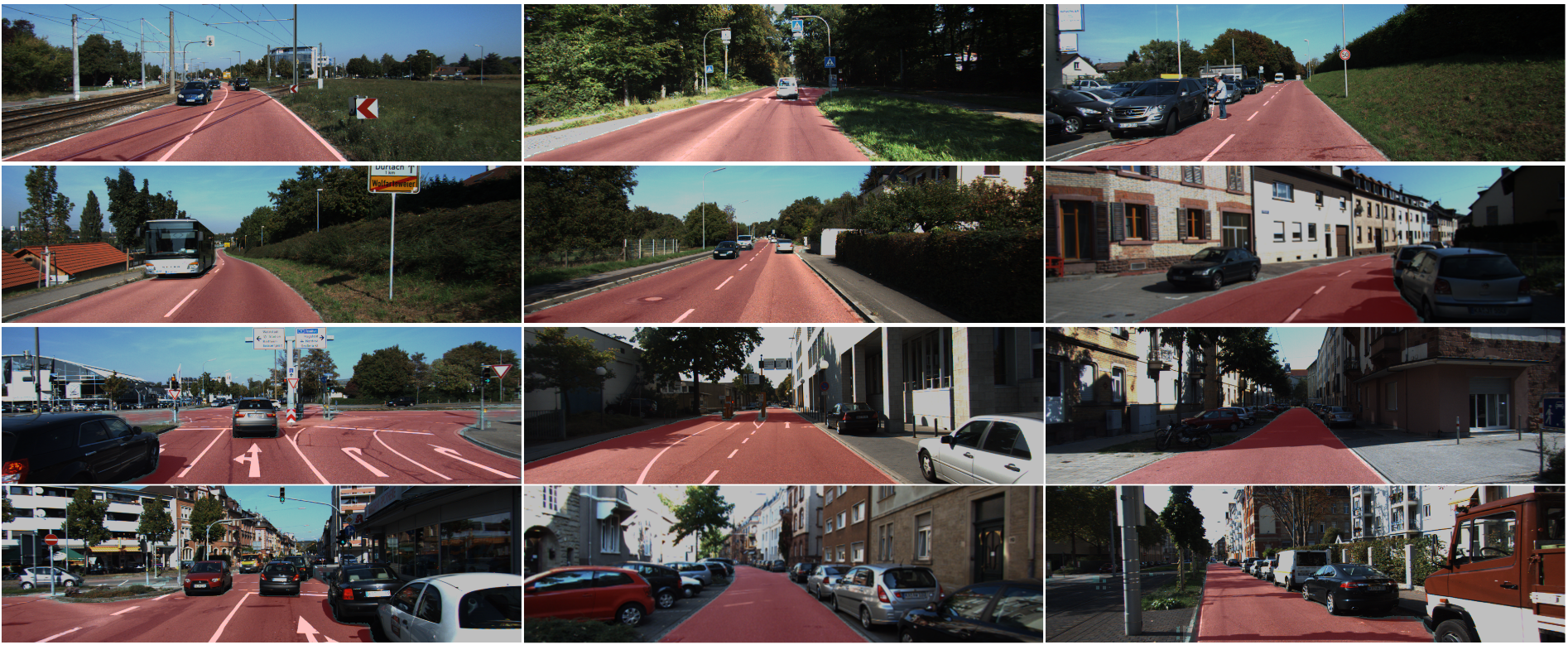}
\end{center}
\caption{Examples of visualization on Kitti road segmentation test set. The road is marked in red.}
\label{Figure6_kitti}
\vspace{-7pt}
\end{figure}

\textbf{Results} We achieve the state-of-the-art results at the time of submission without using any additional information of stereo, laser points and GPS. Specifically, our model attains the highest maximum F1-measure in the sub-categories of  urban unmarked ({UU\_ROAD}), urban multiple marked ({UMM\_ROAD}) and the overall category {URBAN\_ROAD} of all sub-categories, the highest average precision across all three sub-categories and the overall category by the time of submission of this paper. Examples of visualization results are shown in Figure~\ref{Figure6_kitti}. The detailed results are displayed in Table~\ref{tab:kitti} \footnote{For thorough comparison with other methods, please check http://www.cvlibs.net/datasets/kitti/eval\_road.php.}.

\begin{table}[ht]
\begin{center}
\begin{tabular}{c|cc} \toprule[1.5pt]
    & MaxF & AP\\ \midrule
  {UM\_ROAD} & 95.64\% &	93.50\% \\
  {UMM\_ROAD} & 97.62\% & 95.53\% \\
  {UM\_ROAD} & 95.17\% &	92.73\% \\ \midrule[1.25pt]
  {URBAN\_ROAD} & 96.41\% & 93.88\% \\ \bottomrule[1.5pt]
\end{tabular}
\vspace{2pt}
\caption{Performance on different road scenes in KITTI test set. MaxF: Maximum F1-measure, AP: Average precision.}
\label{tab:kitti} 
\end{center}
\vspace{-15pt}
\end{table}

\subsection{PASCAL VOC2012 dataset}
\textbf{Dataset} The PASCAL VOC2012 segmentation benchmark contains $1464$ training images, $1449$ validation images, and $1456$ test images. Using the extra annotations provided by \cite{hariharan2011semantic}, the training set is augmented to have $10582$ images. The dataset has 20 foreground object categories and 1 background class with pixel-level annotation.

\begin{figure}[h]
\begin{center}
\includegraphics[width=0.45\textwidth]{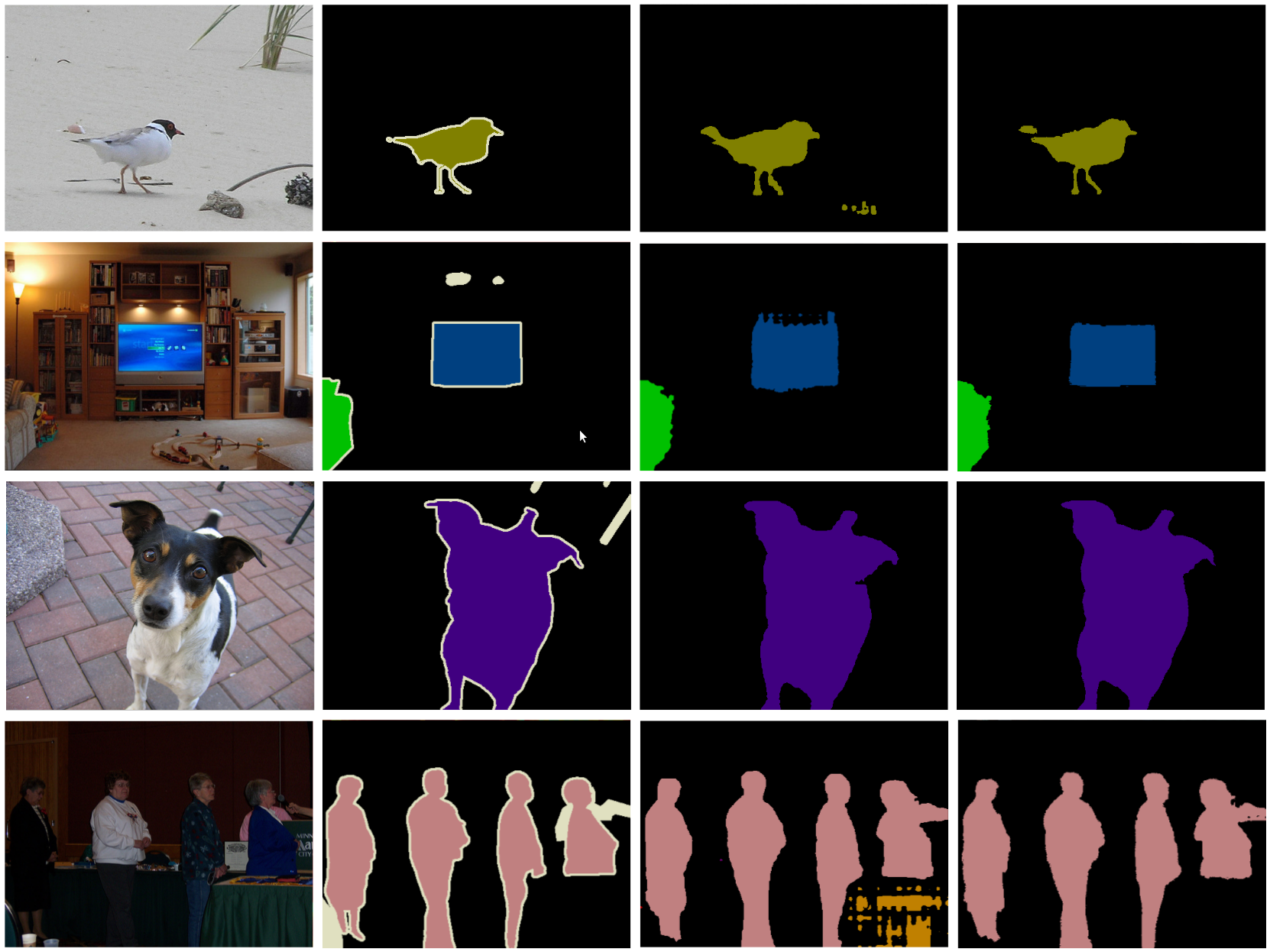}
\end{center}
\caption{Examples of visualization on the PASCAL VOC2012 segmentation validation set. Left to right: input image, ground truth, our result before CRF, and after CRF.}
\label{Figure7_voc}
\vspace{-5pt}
\end{figure}

\textbf{Results} We first pretrain our 152 layer ResNet-DUC model using a combination of augmented VOC2012 training set and MS-COCO dataset \cite{lin2014microsoft}, and then finetune the pretrained network using augmented VOC2012 trainval set. We use patch size $512\times 512$ (zero-padded) throughout training. All other training strategies are the same as Cityscapes experiment. We achieve mIOU of $\textbf{83.1\%}$ on the test set using a single model without any model ensemble or multiscale testing, which is the best-performing method at the time of submission\footnote{Result link: http://host.robots.ox.ac.uk:8080/anonymous/LQ2ACW.html}. The detailed results are displayed in Table~\ref{tab:voc}, and the visualizations are shown in Figure~\ref{Figure7_voc}. 

\begin{table}[ht]
\begin{center}
\begin{tabular}{c|c} \toprule[1.5pt]
    Method & mIoU\\ \midrule
  DeepLabv2-CRF\cite{chen2016deeplab} &	$79.7\%$ \\
  CentraleSupelec Deep G-CRF\cite{chandra2016fast} &	$80.2\%$ \\
  \textbf{ResNet-DUC (ours)} & $\textbf{83.1\%}$ \\ \bottomrule[1.5pt]
\end{tabular}
\vspace{2pt}
\caption{Performance on the Pascal VOC2012 test set.}
\label{tab:voc} 
\end{center}
\vspace{-15pt}
\end{table}

\section{Conclusion}
We propose simple yet effective convolutional operations for improving semantic segmentation systems. We designed a new dense upsampling convolution (DUC) operation to enable pixel-level prediction on feature maps, and hybrid dilated convolution (HDC) to solve the gridding problem, effectively enlarging the receptive fields of the network. Experimental results demonstrate the effectiveness of our framework on various semantic segmentation tasks.

\section{Acknowledgments}
We thank the members of TuSimple and Gary's Unbelievable Research Unit (GURU) for comments on this work. GWC was supported in part by Guangzhou Science and Technology Planning Project (201704030051) and NSF cooperative agreement SMA 1041755 to the Temporal Dynamics of Learning Center, an NSF Science of Learning Center.

{\small
\bibliographystyle{ieee}
\bibliography{wacv_2018}

\begin{thebibliography}{10}\itemsep=-1pt

\bibitem{arnab2015higher}
A.~Arnab, S.~Jayasumana, S.~Zheng, and P.~Torr.
\newblock Higher order potentials in end-to-end trainable conditional random
  fields.
\newblock {\em arXiv preprint arXiv:1511.08119}, 2015.

\bibitem{chandra2016fast}
S.~Chandra and I.~Kokkinos.
\newblock Fast, exact and multi-scale inference for semantic image segmentation
  with deep gaussian crfs.
\newblock {\em arXiv preprint arXiv:1603.08358}, 2016.

\bibitem{chen2016deeplab}
L.-C. Chen, G.~Papandreou, I.~Kokkinos, K.~Murphy, and A.~L. Yuille.
\newblock Deeplab: Semantic image segmentation with deep convolutional nets,
  atrous convolution, and fully connected crfs.
\newblock {\em arXiv preprint arXiv:1606.00915}, 2016.

\bibitem{chen2015mxnet}
T.~Chen, M.~Li, Y.~Li, M.~Lin, N.~Wang, M.~Wang, T.~Xiao, B.~Xu, C.~Zhang, and
  Z.~Zhang.
\newblock Mxnet: A flexible and efficient machine learning library for
  heterogeneous distributed systems.
\newblock {\em arXiv preprint arXiv:1512.01274}, 2015.

\bibitem{Cordts2016Cityscapes}
M.~Cordts, M.~Omran, S.~Ramos, T.~Rehfeld, M.~Enzweiler, R.~Benenson,
  U.~Franke, S.~Roth, and B.~Schiele.
\newblock The cityscapes dataset for semantic urban scene understanding.
\newblock In {\em Proc. of the IEEE Conference on Computer Vision and Pattern
  Recognition (CVPR)}, 2016.

\bibitem{dai2016r}
J.~Dai, Y.~Li, K.~He, and J.~Sun.
\newblock R-fcn: Object detection via region-based fully convolutional
  networks.
\newblock {\em arXiv preprint arXiv:1605.06409}, 2016.

\bibitem{dosovitskiy2015learning}
A.~Dosovitskiy, J.~Tobias~Springenberg, and T.~Brox.
\newblock Learning to generate chairs with convolutional neural networks.
\newblock In {\em Proceedings of the IEEE Conference on Computer Vision and
  Pattern Recognition}, pages 1538--1546, 2015.

\bibitem{pascal-voc-2012}
M.~Everingham, L.~Van~Gool, C.~K.~I. Williams, J.~Winn, and A.~Zisserman.
\newblock The {PASCAL} {V}isual {O}bject {C}lasses {C}hallenge 2012 {(VOC2012)}
  {R}esults.
\newblock
  http://www.pascal-network.org/challenges/VOC/voc2012/workshop/index.html.

\bibitem{fischer2015flownet}
P.~Fischer, A.~Dosovitskiy, E.~Ilg, P.~H{\"a}usser, C.~Haz{\i}rba{\c{s}},
  V.~Golkov, P.~van~der Smagt, D.~Cremers, and T.~Brox.
\newblock Flownet: Learning optical flow with convolutional networks.
\newblock {\em arXiv preprint arXiv:1504.06852}, 2015.

\bibitem{Fritsch2013ITSC}
J.~Fritsch, T.~Kuehnl, and A.~Geiger.
\newblock A new performance measure and evaluation benchmark for road detection
  algorithms.
\newblock In {\em International Conference on Intelligent Transportation
  Systems (ITSC)}, 2013.

\bibitem{ghiasi2016laplacian}
G.~Ghiasi and C.~Fowlkes.
\newblock Laplacian reconstruction and refinement for semantic segmentation.
\newblock {\em arXiv preprint arXiv:1605.02264}, 2016.

\bibitem{hariharan2011semantic}
B.~Hariharan, P.~Arbel{\'a}ez, L.~Bourdev, S.~Maji, and J.~Malik.
\newblock Semantic contours from inverse detectors.
\newblock In {\em 2011 International Conference on Computer Vision}, pages
  991--998. IEEE, 2011.

\bibitem{he2015deep}
K.~He, X.~Zhang, S.~Ren, and J.~Sun.
\newblock Deep residual learning for image recognition.
\newblock {\em arXiv preprint arXiv:1512.03385}, 2015.

\bibitem{holschneider1990real}
M.~Holschneider, R.~Kronland-Martinet, J.~Morlet, and P.~Tchamitchian.
\newblock A real-time algorithm for signal analysis with the help of the
  wavelet transform.
\newblock In {\em Wavelets}, pages 286--297. Springer, 1990.

\bibitem{kokkinos2015pushing}
I.~Kokkinos.
\newblock Pushing the boundaries of boundary detection using deep learning.
\newblock {\em arXiv preprint arXiv:1511.07386}, 2015.

\bibitem{koltun2011efficient}
P.~Kr\"ahenb\"uhl and V.~Koltun.
\newblock Efficient inference in fully connected crfs with gaussian edge
  potentials.
\newblock In {\em Advances in neural information processing systems}, 2011.

\bibitem{krizhevsky2012imagenet}
A.~Krizhevsky, I.~Sutskever, and G.~E. Hinton.
\newblock Imagenet classification with deep convolutional neural networks.
\newblock In {\em Advances in neural information processing systems}, pages
  1097--1105, 2012.

\bibitem{lin2015efficient}
G.~Lin, C.~Shen, I.~Reid, et~al.
\newblock Efficient piecewise training of deep structured models for semantic
  segmentation.
\newblock {\em arXiv preprint arXiv:1504.01013}, 2015.

\bibitem{lin2014microsoft}
T.-Y. Lin, M.~Maire, S.~Belongie, J.~Hays, P.~Perona, D.~Ramanan,
  P.~Doll{\'a}r, and C.~L. Zitnick.
\newblock Microsoft coco: Common objects in context.
\newblock In {\em European Conference on Computer Vision}, pages 740--755.
  Springer, 2014.

\bibitem{liu2015semantic}
Z.~Liu, X.~Li, P.~Luo, C.-C. Loy, and X.~Tang.
\newblock Semantic image segmentation via deep parsing network.
\newblock In {\em Proceedings of the IEEE International Conference on Computer
  Vision}, pages 1377--1385, 2015.

\bibitem{long2015fully}
J.~Long, E.~Shelhamer, and T.~Darrell.
\newblock Fully convolutional networks for semantic segmentation.
\newblock In {\em Proceedings of the IEEE Conference on Computer Vision and
  Pattern Recognition}, pages 3431--3440, 2015.

\bibitem{noh2015learning}
H.~Noh, S.~Hong, and B.~Han.
\newblock Learning deconvolution network for semantic segmentation.
\newblock In {\em Proceedings of the IEEE International Conference on Computer
  Vision}, pages 1520--1528, 2015.

\bibitem{russakovsky2015imagenet}
O.~Russakovsky, J.~Deng, H.~Su, J.~Krause, S.~Satheesh, S.~Ma, Z.~Huang,
  A.~Karpathy, A.~Khosla, M.~Bernstein, et~al.
\newblock Imagenet large scale visual recognition challenge.
\newblock {\em International Journal of Computer Vision}, 115(3):211--252,
  2015.

\bibitem{sevilla2016optical}
L.~Sevilla-Lara, D.~Sun, V.~Jampani, and M.~J. Black.
\newblock Optical flow with semantic segmentation and localized layers.
\newblock {\em arXiv preprint arXiv:1603.03911}, 2016.

\bibitem{shi2016real}
W.~Shi, J.~Caballero, F.~Husz{\'a}r, J.~Totz, A.~P. Aitken, R.~Bishop,
  D.~Rueckert, and Z.~Wang.
\newblock Real-time single image and video super-resolution using an efficient
  sub-pixel convolutional neural network.
\newblock In {\em Proceedings of the IEEE Conference on Computer Vision and
  Pattern Recognition}, pages 1874--1883, 2016.

\bibitem{simonyan2014very}
K.~Simonyan and A.~Zisserman.
\newblock Very deep convolutional networks for large-scale image recognition.
\newblock {\em arXiv preprint arXiv:1409.1556}, 2014.

\bibitem{wavenet}
A.~van~den Oord, S.~Dieleman, H.~Zen, K.~Simonyan, O.~Vinyals, A.~Graves,
  N.~Kalchbrenner, A.~Senior, and K.~Kavukcuoglu.
\newblock Wavenet: A generative model for raw audio.
\newblock {\em arXiv preprint arXiv:1609.03499}, 2016.

\bibitem{wu2016high}
Z.~Wu, C.~Shen, and A.~v.~d. Hengel.
\newblock High-performance semantic segmentation using very deep fully
  convolutional networks.
\newblock {\em arXiv preprint arXiv:1604.04339}, 2016.

\bibitem{yu2015multi}
F.~Yu and V.~Koltun.
\newblock Multi-scale context aggregation by dilated convolutions.
\newblock {\em arXiv preprint arXiv:1511.07122}, 2015.

\bibitem{zeiler2014visualizing}
M.~D. Zeiler and R.~Fergus.
\newblock Visualizing and understanding convolutional networks.
\newblock In {\em European Conference on Computer Vision}, pages 818--833.
  Springer, 2014.

\bibitem{zheng2015conditional}
S.~Zheng, S.~Jayasumana, B.~Romera-Paredes, V.~Vineet, Z.~Su, D.~Du, C.~Huang,
  and P.~H. Torr.
\newblock Conditional random fields as recurrent neural networks.
\newblock In {\em Proceedings of the IEEE International Conference on Computer
  Vision}, pages 1529--1537, 2015.

\bibitem{zhou2016semantic}
B.~Zhou, H.~Zhao, X.~Puig, S.~Fidler, A.~Barriuso, and A.~Torralba.
\newblock Semantic understanding of scenes through the ade20k dataset.
\newblock {\em arXiv preprint arXiv:1608.05442}, 2016.

\end{thebibliography}
}

\end{document}